\documentclass[runningheads]{llncs}
\usepackage{amsmath}
\usepackage{array}
\usepackage{booktabs}
\usepackage{multirow}
\usepackage[table]{xcolor}
\usepackage{amssymb}
\usepackage{enumitem}
\usepackage{wrapfig}
\usepackage{colortbl}
\usepackage{dirtytalk}

\usepackage[most]{tcolorbox}
\usepackage[colorlinks=true,linkcolor=caribbeangreen,citecolor=emerald]{hyperref}
\usepackage[colorinlistoftodos,prependcaption,textsize=tiny]{todonotes}
\usepackage[align=center,shadow=true,shadowsize=5pt,nobreak=true,framemethod=tikz,style=0,skipabove=2pt,skipbelow=1pt,innertopmargin=-3pt,innerbottommargin=3pt,innerleftmargin=5pt,innerrightmargin=5pt,leftmargin=-2pt,rightmargin=-2pt]{mdframed}
\definecolor{babyblue}{rgb}{0.54, 0.81, 0.94}
\definecolor{armygreen}{rgb}{0.29, 0.33, 0.13}
\definecolor{brightlavender}{rgb}{0.75, 0.58, 0.89}
\definecolor{aqua}{rgb}{0.0, 1.0, 1.0}
\definecolor{caribbeangreen}{rgb}{0.0, 0.8, 0.6}
\definecolor{reddish}{rgb}{0.82, 0.1, 0.26}
\definecolor{emerald}{rgb}{0.31, 0.78, 0.47}
\definecolor{jasper}{rgb}{0.84, 0.23, 0.24}
\definecolor{red}{rgb}{1.0, 0.0, 0.0}
\definecolor{green}{rgb}{0.0, 1.0, 0.0}
\definecolor{blue}{rgb}{0.0, 0.0, 1.0}
\definecolor{darkgreen}{rgb}{0.1, 0.7, 0.1}
\definecolor{darkblue}{rgb}{0.1, 0.1, 0.7}
\definecolor{red}{rgb}{0.7, 0.1, 0.1}
\usepackage{graphicx,verbatim}
\usepackage{orcidlink}
\usepackage{comment}
\usepackage{tikz}
\usepackage{xparse}   

\DeclareRobustCommand{\authorpic}[2][5mm]{%
  \tikz[baseline={([yshift=-.25ex]current bounding box.center)}]{%
    \clip (0,0) circle (#1);
    \pgfmathsetlengthmacro{\picside}{sqrt(2)*#1}%
    \node at (0,0) {\includegraphics[width=\picside,height=\picside,keepaspectratio]{#2}};
    \draw[line width=0.4pt, color=white] (0,0) circle (#1);
  }%
}

\NewDocumentCommand{\AuthorWithPic}{O{5.5mm} O{0.20em} m m}{%
  \texorpdfstring{\authorpic[#1]{#4}\kern #2}{}%
  #3%
}

\usepackage{xcolor}
\definecolor{linkpinkix}{HTML}{EA335A} 

\definecolor{linkpink}{HTML}{EA335A}

\newcommand{\shadedlink}[2]{%
  \tikz[baseline=(n.base)]\node[
    fill=linkpink,
    fill opacity=0.5,
    text opacity=1,
    rounded corners=.3ex,
    inner xsep=.35em,
    inner ysep=.15em
  ] (n) {\href{#1}{\textcolor{blue!70!black}{#2}}};%
}

\begin{document}
%

\title{X-Node: Self-Explanation is All We Need}

\author{%
  \AuthorWithPic[6mm][0.18em]{Prajit Sengupta}{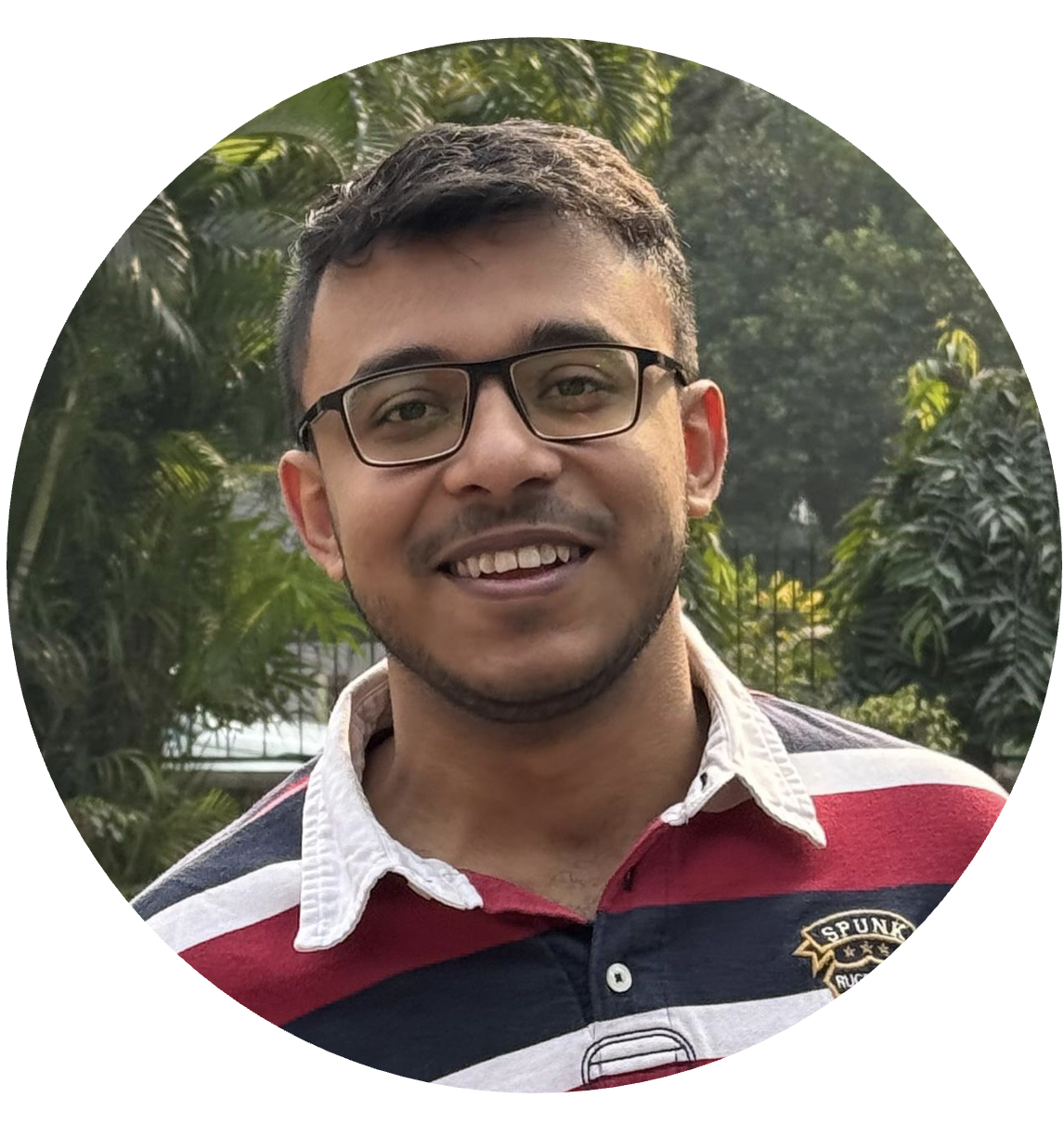}\orcidlink{0009-0006-9930-7847} \and
  \AuthorWithPic[6mm][0.18em]{Islem Rekik}{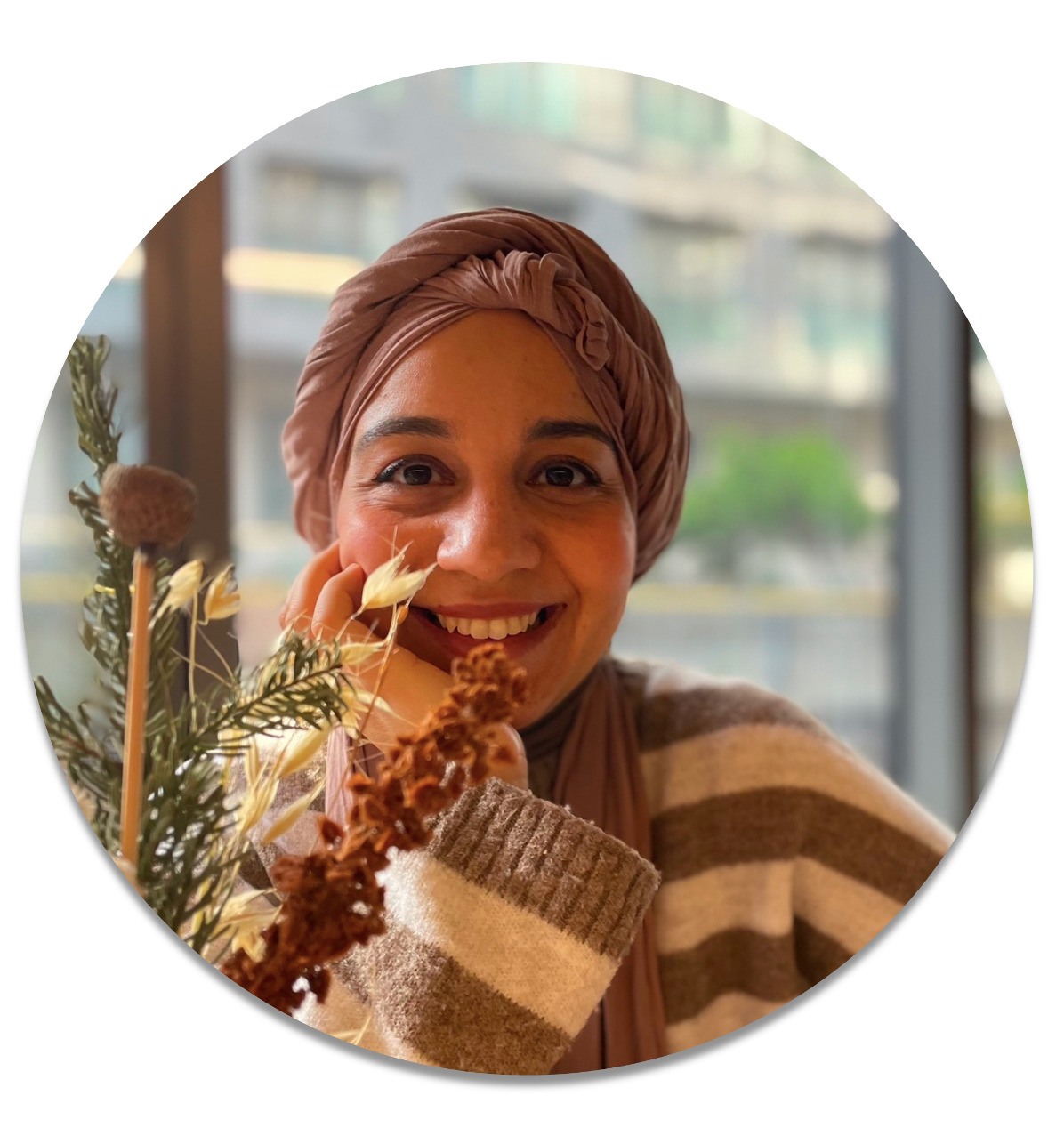}\orcidlink{0000-0001-5595-6673}\thanks{Corresponding author: \email{i.rekik@imperial.ac.uk}, \url{http://basira-lab.com}, GitHub: \url{https://github.com/basiralab/X-Node}}%
}

\institute{BASIRA Lab, Imperial-X (I-X) and Department of Computing, Imperial College London, London, United Kingdom}
\authorrunning{Sengupta et al.}

\maketitle

\begin{abstract}
Graph neural networks (GNNs) have achieved state-of-the-art results in computer vision and medical image classification tasks by capturing structural dependencies across data instances. However, their decision-making remains largely opaque limiting their trustworthiness in high-stakes clinical applications, where interpretability is essential. Existing explainability techniques for GNNs are typically post-hoc and global, offering limited insight into individual node decisions or local reasoning. We introduce \textbf{X-Node}, a self- explaining GNN framework in which each node generates its own explanation as part of the prediction process. For every node, we construct a structured \textit{context vector} encoding interpretable cues, such as degree, centrality, clustering, feature saliency, and label agreement within its local topology. A lightweight \textit{Reasoner} module maps this context into a compact \textit{explanation vector}, which serves three purposes: (1) reconstructing the node’s latent embedding via a Decoder to enforce faithfulness, (2) generating a natural language explanation using a pre-trained LLM (e.g., Grok or Gemini), and (3) guiding the GNN itself via a “text-injection” mechanism that feeds explanations back into the message-passing pipeline. We evaluate \textbf{X-Node} on two graph datasets derived from \textit{MedMNIST} and \textit{MorphoMNIST,} integrating it with GCN, GAT, and GIN backbones. Our results show that X-Node maintains competitive classification accuracy while producing faithful, per-node explanations. \url{https://github.com/basiralab/X-Node}.\footnote{This paper has been selected for an \textbf{Oral Presentation} at the GRAIL MICCAI 2025 workshop. \shadedlink{https://youtu.be/pdUGQr6dnt4}{[X-Node YouTube Video]}.}
\end{abstract}


\section{Introduction}
Graph neural networks (GNNs) have emerged as powerful tools for modeling structured medical data, such as cellular interactions in histopathology, organ topologies in medical imaging, and anatomical relationships in population-level brain graphs \cite{going_beyond_euclidian,disease_pred_gnn,graph_med_diagnosis}. By learning over nodes and edges, GNNs naturally capture both local and global structure in such data, and have shown state-of-the-art performance in various diagnostic tasks \cite{GNN_Neural,gated_Fully_fusion}. 
\begin{figure}
    \centering
    \includegraphics[width=0.9\linewidth]{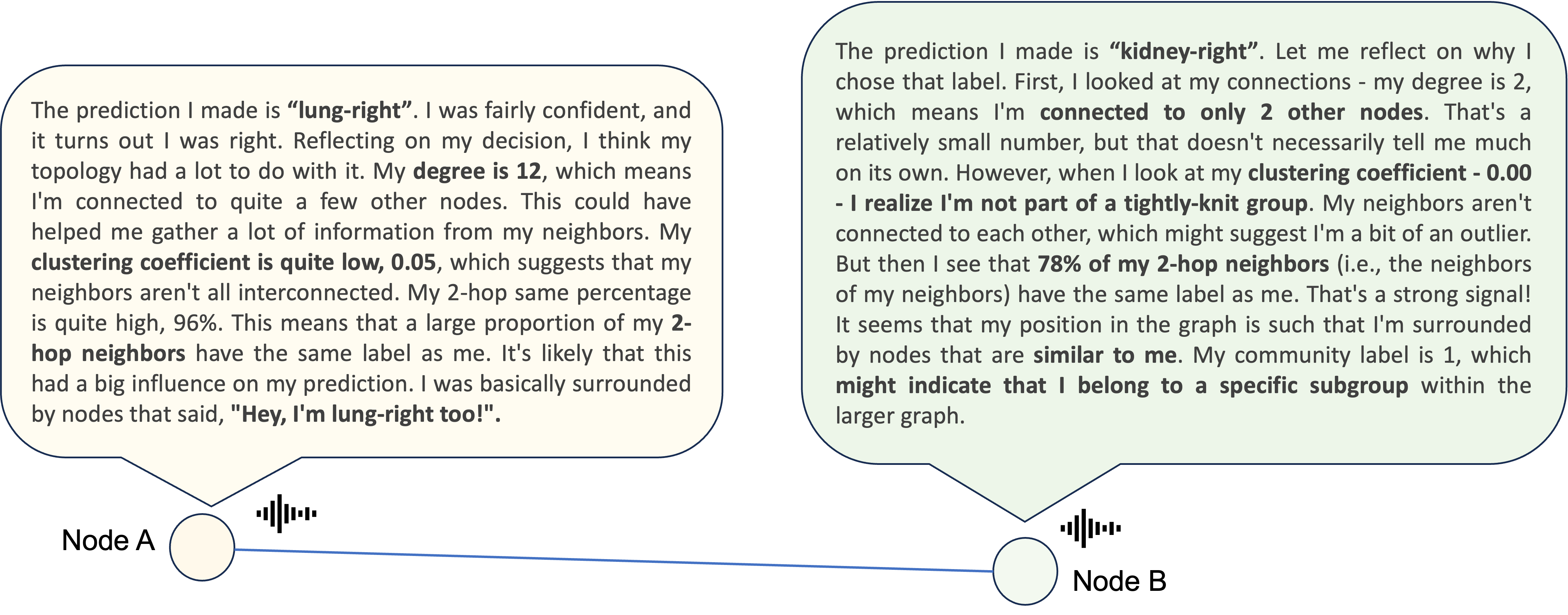}
    \caption{Two neighbouring nodes self-explaining}
    \label{Node.png}
    \vspace{-5mm}
\end{figure}However, \textbf{interpretability remains a key bottleneck}: in high-stakes clinical environments, it is insufficient for models to be merely accurate—they must also explain their decisions in a way that is faithful, transparent, and verifiable \cite{whyXAI_medical_domain,xai_deeplearning}. Despite growing interest in graph explainability \cite{gnn_explainer,parameterized_explainer,XGNN}, current approaches face several critical limitations: \textbf{First}, most existing explainability methods for GNNs are \textit{post-hoc} and \textit{non-intrinsic}. Tools such as GNNExplainer \cite{gnn_explainer} or PGExplainer \cite{PGExplainar} identify subgraphs or features after training, with no guarantee that these explanations reflect the model’s actual reasoning process \cite{LIMESHAP}. Such post-hoc explanations are prone to instability and adversarial inconsistency, especially in sparse medical graphs where subtle topological changes can shift model predictions without warning \cite{post_hoc_fail}. \textbf{Second}, existing models do not offer \textit{localized, node-level reasoning}. A GNN’s decision for a given node emerges from hidden message passing over the graph, but rarely can the node itself articulate why it received a certain label. In contrast, clinical reasoning is often local and explainable: radiologists, for instance, interpret findings based on visual features and regional context. Node-level explainability—where each node reasons about its own state as shown in Fig.~\ref{Node.png}—is notably absent from current architectures. \textbf{Finally}, current GNN pipelines treat explanation as disconnected from learning. That is, even if explanations are available, they are not used to guide or constrain training. This decoupling limits the utility of explanations in improving robustness or trust. Worse, explanations can be optimized separately to “look plausible”, but still diverge from the actual decision pathway—a phenomenon known as \textit{rationalization over reasoning} \cite{faithful_interpretable}. To bridge this gap between decision accuracy and faithful interpretability, we introduce \textbf{X-Node}, a fully novel graph learning framework that equips each node with a self-explainable capability. By modeling nodes as introspective agents—capable of constructing contextual explanations, reflecting on their local topology and feature space, and injecting this explainability reasoning back into the network—\textbf{X-Node} offers an interpretable and performance-aligned solution to node classification as summarized below:
\vspace{-1mm}
\begin{enumerate}
    \item \textit{On a methodological level:} X-Node introduces \textit{self-explainable graph nodes} by embedding explanation-aware learning directly into the GNN. Each node builds a local context, generates an explanation vector, and reinjects it into the network— enabling faithful, intrinsic interpretability beyond post-hoc approaches.
    \item \textit{On a clinical level:} X-Node allows each node—representing a patient, organ, or region—to justify its prediction using topological cues, label agreement, and optimization signals. This supports transparent, clinician-aligned decision-making.
    \item \textit{On a generic level:} Though evaluated on MedMNIST and MorphoMNIST, X-Node can augment any GNN (e.g., GCN, GAT, GIN) with self-explaining capabilities, offering a modular interpretability layer across graph learning tasks.
\end{enumerate}

\section{Related Work}
\begin{figure}
    \centering
    \includegraphics[width=0.8\linewidth]{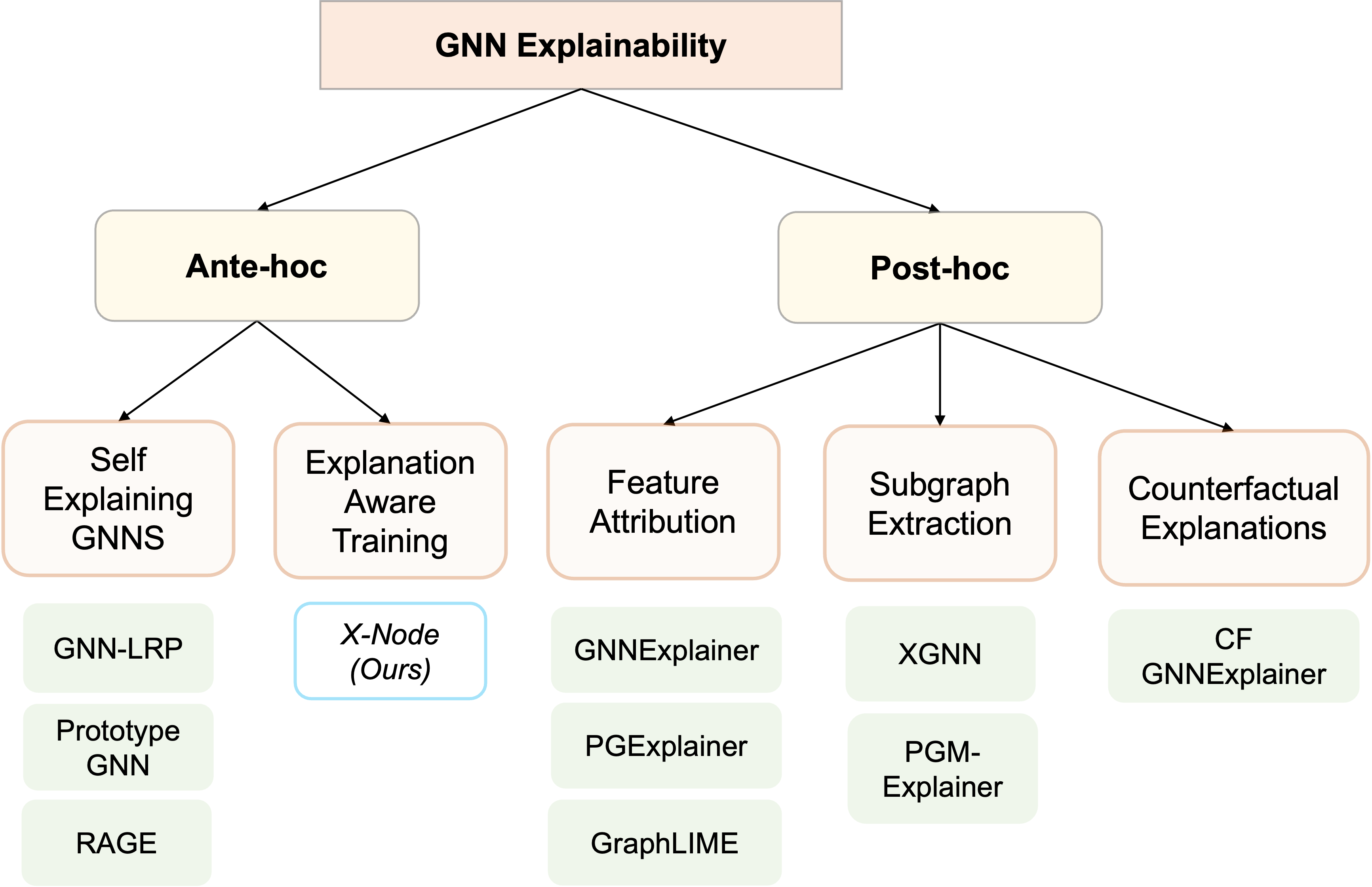}
    \caption{Types of GNN explainability methods}
    \label{fig:gnn_explainability}
    \vspace{-4mm}
\end{figure}

Understanding the landscape of explainability techniques in GNNs is essential for positioning our contribution. Fig.~\ref{fig:gnn_explainability} summarizes this taxonomy, categorizing existing methods as either post-hoc or ante-hoc, and highlighting where our approach, \textbf{X-Node}, fits within this space.

\noindent \textbf{Graph neural networks} Our work builds on the rich literature of GNNs for node classification. Kipf and Welling introduced GCNs, which perform spectral convolutions on graphs \cite{kipf2017semisupervised}. GAT adds attention weights over neighbors \cite{gat}, while \cite{gin_powerfulGNN} proposed GIN (Graph Isomorphism Networks) showed how certain aggregation functions can make GNNs as powerful as the Weisfeiler-Lehman test. We use these classic architectures (GCN, GAT, GIN) as baselines in our experiments. However, unlike standard GNNs, our architecture includes \emph{reasoning modules} that explicitly compute explanations at every node. 

\noindent \textbf{Explainable AI and GNN explanation} Interpretability has long been recognized as essential in critical fields like healthcare \cite{self_explaining_nn}. Many methods seek to explain deep models post-hoc (e.g. LIME, SHAP in vision/NLP), but these do not guarantee faithful introspection \cite{egnn_fire}. In graphs, GNNExplainer \cite{gnn_explainer}
 was among the first general, model-agnostic explainers: it finds a subgraph and feature mask that maximize the mutual information with a given prediction, identifying a compact rationalizing substructure. Other approaches (PGExplainer, XGNN) also produce edge/feature importance masks. While useful, these methods are post-hoc and have been shown to be brittle: even small graph perturbations can drastically change their output without affecting the model’s prediction \cite{egnn_fire}. In contrast, X-Node embeds explainability intrinsically into the model. Our approach is inspired by the framework of self-explaining neural networks, where explanations are generated by the model itself as part of the forward pass.

 \noindent \textbf{Self-explaining models} \cite{self_explaining_nn} introduced SENN, a class of networks where predictions come with explicit contributions and concepts. Similarly, modular or capsule networks use interpretable sub-components. Our Self-Explainable Nodes generalize this idea to graph domains: each node uses its local graph-based context to self-reason about its label. This is akin to building a hybrid deep learning with symbolic reasoning at the node level. The use of language models for explanations also resonates with recent work on chain-of-thought reasoning in LLMs
\cite{chainthought_llm}. 
\begin{table}[h]
\vspace{-3mm}
\centering
\caption{Limitations of current GNN-based medical pipelines and proposed solutions.}
\renewcommand{\arraystretch}{1.5}
\scalebox{0.80}{
\begin{tabular}{>{\raggedright\arraybackslash}p{6.4cm} >{\raggedright\arraybackslash}p{7.6cm}}
\rowcolor{gray!15}
\textbf{Limitations} & \textbf{Proposed Solutions} \\
\midrule

\rowcolor{gray!5}
\textit{Opaque predictions:} Standard GNNs yield high accuracy but provide little insight into \textit{why} a decision was made. In medicine this lack of interpretability reduces clinician trust \cite{gnn_explainer,xai_deeplearning,whyXAI_medical_domain}. 
&
\textbf{Self-Explainable Nodes:} Each node produces an explanation vector reflecting explicit reasoning (e.g., graph topology and 2-hop label patterns). This makes the model’s logic transparent. 
\\

\rowcolor{gray!10}
\textit{Post-hoc explanations can mislead:} Common explainers (e.g., Grad-CAM, LIME) are post-hoc and may not reflect the model's true reasoning. Clinicians cannot rely on them in high-stakes cases \cite{gnn_explainer,LIMESHAP,egnn_fire}. 
&
\textbf{Integrated Explainability:} We incorporate explanation into the training loop (self-explainable AI). Because the Reasoner is part of the model, its explanations are inherently aligned with decision-making, improving faithfulness by aligning symbolic and neural features. 
\\

\rowcolor{gray!5}
\textit{No feedback from explanations:} Conventional pipelines do not use explanations to improve learning. The GNN and the explainer are independent, so explanatory signals do not shape the model \cite{faithful_interpretable,self_explaining_nn}. 
&
\textbf{Feedback Loop:} The explanation embeddings are \textit{injected} back into the GNN layers. This adaptive reinjection lets the model refine its representations based on its own reasoning, effectively using explanation loss as auxiliary supervision. 
\\
\bottomrule
\end{tabular}}
\vspace{-3mm}
\label{tab:limitations-solutions}
\end{table}

\noindent \textbf{LLM-Augmented graph explanation models} Recent advancements have explored the integration of Large Language Models (LLMs) to enhance the interpretability of Graph Neural Networks (GNNs). For instance, GraphXAIN \cite{Gxain} employs LLMs to translate technical outputs, such as subgraphs and feature importance scores, into coherent natural language narratives, thereby improving the understandability of GNN predictions for non-expert users. Similarly, LLMExplainer \cite{LLMExplainer} integrates LLMs as Bayesian inference modules within GNN explanation networks to mitigate learning biases and generate more robust explanations. Our approach diverges from these by utilizing an LLM as an explanation decoder within the GNN framework. In our model, each node constructs a local context vector, which is then transformed into a natural language explanation by the LLM. This explanation is subsequently reintegrated into the GNN through a process we term \say{text-injection}, guiding further message passing and enhancing interpretability. Unlike methods that rely on symbolic reasoning components, our framework leverages the generative capabilities of LLMs to produce human-readable justifications. Throughout this paper, we are formulating \textbf{3 hypotheses}:

\noindent \textbf{H1: Node-level explainability improves interpretability} \textit{Each node can produce a faithful, self-contained explanation by summarizing local graph structure and topology.}

\noindent \textbf{H2: Reasoning as regularization improves learning} \textit{Injecting explanation vectors into GNNs provides inductive bias, improving generalization and aligning embeddings.}

\noindent\textbf{H3: LLMs enhance explanation} \textit{A pre-trained LLM can map structured context into fluent, human-readable text that captures reasoning aligned with clinical logic.}

\section{Methodology}
\begin{figure}
\vspace{-5mm}
    \centering
    \includegraphics[width=\linewidth]{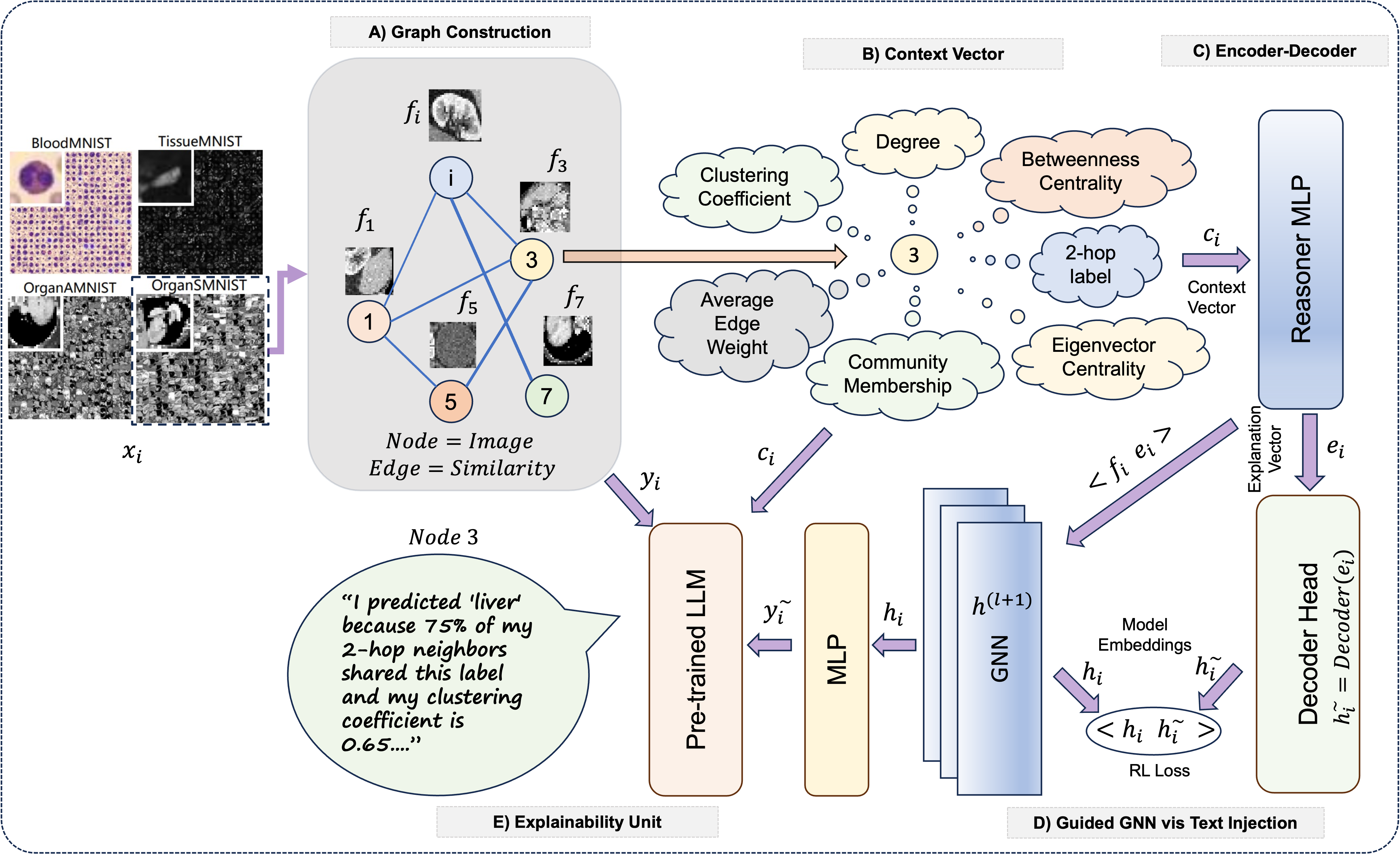}
    \caption{Proposed X-Node Architecture}
    \label{fig:niyan}
    \vspace{-5mm}
\end{figure}

Our proposed framework, \textbf{X-Node}, introduces self-explainable nodes that classify and explain decisions based on graph structure, features, and topological characteristics during training. The overall pipeline is shown in Figure~\ref{fig:niyan}.

\noindent \textbf{A. Problem setup and graph construction}
Let $\mathcal{D} = {(x_i, y_i)}_{i=1}^N$ denote a dataset of $N$ medical images with class labels $y_i \in \mathcal{Y}$. Each image $x_i$ is encoded into a feature vector $f_i \in \mathbb{R}^d$ using a pre-trained CNN encoder $F$, i.e., $f_i = F(x_i)$. We then build a $k$-nearest neighbor ($k$-NN) graph $\mathcal{G} = (\mathcal{V}, \mathcal{E}, \mathbf{X})$, $\mathcal{V}$: Nodes representing images.
$\mathbf{X} = [f_1, \dots, f_N]^\top \in \mathbb{R}^{N \times d}$: Node features.
$\mathcal{E}$: Edges based on cosine similarity encoded into the adjacency matrix:
$A_{ij} = \cos(f_i, f_j) \ \text{if} \ f_j \in \text{Top-}k(f_i); \ 0 \ \text{otherwise}$. This graph encodes both visual and structural similarity across instances as in (Fig~\ref{fig:niyan}-A). 

\noindent \textbf{B. Context vector extraction from graph topology}
In graph-based medical datasets, a node’s local topology often encodes clinically relevant patterns—such as label homophily (e.g., organs of the same type clustering together), centrality (e.g., typical vs. atypical organs), or community structure as shown in (Fig.~\ref{fig:niyan}-B). To capture such interpretable patterns, we construct a compact \textit{context vector} $\mathbf{c}_i \in \mathbb{R}^{d_c}$ for each node $v_i$ based on topological and label-aware descriptors. These serve as reasoning cues for explanation generation and auxiliary supervision. While this vector is numerical for model input, it is preserved as labeled key-value pairs (e.g. \say{degree}: 3, \say{2-hop agreement}: 0.85) where each key is an interpretable feature.
\begin{equation}
\mathbf{c}_i = \text{Concat}(d_i, cc_i, \rho_i^{(2)}, ec_i, bc_i, \bar{w}_i, c_i) 
\end{equation}
The following features are selected for their empirical and theoretical relevance to graph classification, especially in sparse or hierarchical domains like medical imaging:
\begin{center}
\vspace{-0.5mm}
\scalebox{0.7}{
\begin{tcolorbox}[colback=green!5!white, colframe=green!40!black, width=0.95\linewidth, boxrule=0.5pt, arc=4pt]
\vspace{-2mm}
\begin{multicols}{2}
\begin{itemize}[leftmargin=*, itemsep=0.6em]
    \item \textbf{Degree} ($d_i$): Node connectivity. High degree may indicate prototypical or redundant nodes.
    
    \item \textbf{Clustering Coefficient} ($cc_i$): Reflects local cohesion. High values suggest dense neighborhoods.
    
    \item \textbf{2-hop Label Agreement} ($\rho_i^{(2)}$): Measures semantic consistency in the extended neighborhood:
    \[
    \rho_i^{(2)} = \frac{\# \text{ same-label nodes in 2-hop}}{\# \text{ 2-hop neighbors}}
    \]
    
    \item \textbf{Eigenvector Centrality} ($ec_i$): Importance in global graph flow. Nodes connected to other important nodes receive high scores.
    
    \item \textbf{Betweenness Centrality} ($bc_i$): Captures "bridge" roles. Important for detecting outliers or misclassified nodes.
    
    \item \textbf{Average Edge Weight} ($\bar{w}_i$): Indicates confidence in neighborhood similarity:
    \[
    \bar{w}_i = \frac{1}{|\mathcal{N}(i)|} \sum_{j \in \mathcal{N}(i)} w_{ij}
    \]
    
    \item \textbf{Community Membership} ($c_i$): Structural cluster ID, indicating coarse graph-level partitioning.
\end{itemize}
\end{multicols}
\end{tcolorbox}}
\vspace{-1mm}
\end{center}

\noindent \textbf{C. Explanation vector generation via Reasoner}  
To interpret a node’s behavior, its context vector $\mathbf{c}_i$ is passed through a MLP Reasoner parameterized by trainable weights $W_1, W_2$ and biases $b_1, b_2$ to produce a low-dimensional explanation vector $\mathbf{e}_i \in \mathbb{R}^{d_e}$:  
\begin{equation}
\mathbf{e}_i = \text{Reasoner}\,\phi(\mathbf{c}_i) = \sigma(W_2 \cdot \text{ReLU}(W_1 \cdot \mathbf{c}_i + b_1) + b_2)
\end{equation}

\noindent \textbf{D. Embedding reconstruction via decoder}
To ensure faithfulness, $\mathbf{e}_i$ is decoded to reconstruct the node’s latent GNN embedding $\hat{\mathbf{h}}_i \in \mathbb{R}^{d_h}$:  
$\hat{\mathbf{h}}_i = \text{Decoder}\,\psi(\mathbf{e}_i)$ (Fig.~\ref{fig:niyan}-C). This enforces alignment between $\hat{\mathbf{h}}_i$ and the actual embedding $\mathbf{h}_i$ from the GNN.

\noindent \textbf{E. Textual explanation via LLM}
To generate interpretable explanations (Fig.~\ref{fig:niyan}-E), each node $v_i$ uses its structured \textit{context vector} $\mathbf{c}_i \in \mathbb{R}^{d_c}$ and its predicted label $\hat{y}_i$ (and optionally, true label $y_i$) as input to a pre-trained large language model (LLM), such as Grok's \texttt{llama-4-scout-17b-16e-instruct} or Google's Gemini 2.5 Pro:
\begin{equation}
\mathcal{T}_i = \text{LLM}_\psi\left(\texttt{prompt}(\mathbf{c}_i, \hat{y}_i, y_i)\right)
\end{equation}
\noindent The LLM serves as a natural language decoder, converting structured node-level statistics into faithful textual rationales $\mathcal{T}_i$ supporting \textbf{H3}. The prompt is formatted as follows:

\begin{tcolorbox}[colback=gray!5!white, colframe=black!40, title=LLM Prompt for Explanation]
\vspace{-2mm}
\small
You are a node in a medical graph.\\
Your topological context is: \texttt{<context\_vector>}\\
Your predicted label: \texttt{<predicted\_label>}. True label: \texttt{<true\_label>} \\
Explain in natural language why you predicted \texttt{<predicted\_label>}. If incorrect, describe what might have misled you based on your structure, features, and neighbors.
\vspace{-2mm}
\end{tcolorbox}
\vspace{-0.5mm}
\noindent This design allows the node to \textit{self-narrate} its decision logic—either validating a correct classification or introspecting its own failure. 

\noindent \textbf{F. Explanation-guided GNN via text injection} 
The explanation vector $\mathbf{e}_i$ is concatenated with the GNN embedding $\mathbf{h}_i$ for final classification, allowing reasoning signals to directly inform prediction, creating a feedback loop as shown in (Fig.~\ref{fig:niyan}-D): 
\begin{equation}
\mathbf{z}_i = \text{Concat}(\mathbf{h}_i, \mathbf{e}_i), \quad \hat{y}_i = \text{MLP}{\text{class}}(\mathbf{z}_i)
\end{equation}
\noindent \textbf{G. Loss function and joint training}
The model is trained by jointly minimizing classification, alignment, and reconstruction losses:
\begin{equation}
\mathcal{L} = \sum_{i=1}^N \left[
\underbrace{\text{CE}(\hat{y}_i, y_i)}_{\text{Classification}} +
\alpha \underbrace{\|\mathbf{e}_i - \mathbf{c}_i\|^2}_{\text{Alignment}} +
\beta \underbrace{\|\hat{\mathbf{h}}_i - \mathbf{h}_i\|^2}_{\text{Reconstruction}}
\right]
\end{equation}

\section{Results and Discussion}

We assess X-Node on six image-derived graph datasets—five from MedMNIST (OrganCMNIST, OrganAMNIST, OrganSMNIST, TissueMNIST, BloodMNIST) \cite{yang2023medmnist} and one synthetic benchmark (Morpho-MNIST) \cite{moprhomnist_data}. Each dataset is converted into a $k$-NN graph using pretrained image embeddings. X-Node is evaluated not only for classification accuracy but also for its ability to generate faithful per-node explanations. Results are averaged over \textbf{3-fold cross-validation} with seeds 42, 43, and 44 (9 experiments).

\begin{table}[h]
\vspace{-5mm}
\centering
\caption{Graph Dataset Details created from Images}
\label{datasets}
\scalebox{0.67}{
\begin{tabular}{|l|c|c|c|c|c|c|}
\hline
 & \textbf{OrganCMNIST} & \textbf{OrganAMNIST} & \textbf{OrganSMNIST} & \textbf{TissueMNIST} & \textbf{BloodMNIST} & \textbf{Morpho-MNIST} \\
\hline
\# of Nodes & 23583 & 58830 & 25211 & 236386 & 17092 & 280000 \\
\# of Edges & 82315 & 205404 & 88951 & 826714 & 60116 & 970699 \\
\# of Features & 512 & 512 & 512 & 512 & 512 & 512 \\
\# of Labels & 11 & 11 & 11 & 8 & 8 & 4 \\
Task Type & Multi-class & Multi-class & Multi-class & Multi-class & Multi-class & Multi-class \\
Training Type & Inductive & Inductive & Inductive & Inductive & Inductive & Inductive \\
\hline
Training Nodes & 12975 & 34561 & 13932 & 165466 & 11959 & 216000 \\
Validation Nodes & 2392 & 6491 & 2452 & 23640 & 1712 & 24000 \\
Test Nodes & 8216 & 17778 & 8827 & 47280 & 3421 & 40000 \\
\hline
\end{tabular}}
\vspace{-6mm}
\end{table}

\noindent \textbf{Dataset \& experimentation} As shown in Table \ref{datasets} the datasets span a range of medical domains, with node counts from 17K (BloodMNIST) to 236K (TissueMNIST), and label spaces from 4 to 11 classes. Experiments ran on an Apple M2 Air with 16GB RAM and MPS acceleration GPU, using an 80/20 train-validation split in a 3-fold CV setup. Adding \textit{reasoner} increased epoch time and memory moderately as in Table \ref{tab:efficiency1}.
\vspace{-6mm}
\begin{table}[h!]
\centering
\caption{Comparison of \textbf{Average Epoch Time (s)} and \textbf{Peak Memory (MB)} across datasets.}
\label{tab:efficiency1}
\resizebox{\textwidth}{!}{%
\begin{tabular}{|l|cc|cc|cc|cc|cc|cc|}
\hline
\textbf{Method} & \multicolumn{2}{c|}{\textbf{OrganCMNIST}} & \multicolumn{2}{c|}{\textbf{OrganAMNIST}} & \multicolumn{2}{c|}{\textbf{OrganSMNIST}} & \multicolumn{2}{c|}{\textbf{TissueMNIST}} & \multicolumn{2}{c|}{\textbf{BloodMNIST}} & \multicolumn{2}{c|}{\textbf{MoprhoMNIST}} \\
 & \textbf{Time} & \textbf{Memory} & \textbf{Time} & \textbf{Memory} & \textbf{Time} & \textbf{Memory} & \textbf{Time} & \textbf{Memory} & \textbf{Time} & \textbf{Memory} & \textbf{Time} & \textbf{Memory} \\
\hline
GCN & 0.48 & 758.95 & 1.15 & 1234.64 & 0.49 & 1024.92 & 2.76 & 2472.47 & 0.35 & 557.36 & 4.20 & 2648.61 \\
GCN + Reasoner & 1.90 & 944.25 & 1.78 & 1439.06 & 1.25 & 889.30 &  4.76 & 1781.39 & 0.38 & 718.44 & 4.71 & 1717.73 \\
\hline
\end{tabular}%
}
\end{table}
\vspace{-6mm}

\begin{table}[ht]
\label{gnn_varaints}
\vspace{-1mm}
\centering
\scriptsize  
\setlength{\tabcolsep}{2.0pt}
\renewcommand{\arraystretch}{1.2}
\caption{Comparison of GCN variants on six inductive datasets (best results in green).}
\label{tab:gcnsixdatasets}

  \scalebox{0.9}{\begin{tabular}{l
  *{2}{>{\centering\arraybackslash}p{1.3cm}
        >{\centering\arraybackslash}p{1.3cm}
        >{\centering\arraybackslash}p{1.3cm}
        >{\centering\arraybackslash}p{1.3cm}}}
\toprule
& \multicolumn{4}{c}{\bfseries OrganCMNIST}
& \multicolumn{4}{c}{\bfseries OrganAMNIST} \\
\cmidrule(lr){2-5}\cmidrule(lr){6-9}
\bfseries Method
  & ACC & F1   & Sensitivity  & ROC-AUC
  & ACC & F1   & Sensitivity  & ROC-AUC \\
\midrule
GCN & 88.20±0.61 & 86.03±0.89 & 86.13±0.06 & 99.09±0.08
    & 91.85±0.30 & 91.19±0.33 & 91.18±0.03 & 99.51±0.03\\

\rowcolor{green!40}
GCN + Reasoner & 89.22±0.79 & 87.46±0.87 & 87.93±0.08 & 99.18±0.07 & 93.64±0.21 & 93.16±0.19 & 93.36±0.02 & 99.64±0.03 \\
\arrayrulecolor{gray}\hline
GAT & 90.31±0.28 & 88.47±0.34 & 88.51±0.03 & 99.38±0.05 & 93.69±0.36 & 93.26±0.28 & 93.36±0.04 & 99.69±0.02
\\
\rowcolor{green!40}
GAT+ Reasoner & 90.75±0.54 & 89.11±0.60 & 89.29±0.55 & 99.37±0.05 & 94.17±0.20 & 93.85±0.16 & 94.07±0.13 & 99.69±0.02
\\
\arrayrulecolor{gray}\hline
GIN & 87.96±0.59 & 85.61±0.75 & 85.56±0.75 & 98.86±0.09 & 91.54±0.71 & 90.45±0.73 & 90.50±0.69 & 99.41±0.08 \\
\rowcolor{green!40}
GIN + Reasoner & 89.61±0.24 & 87.78±0.24 & 87.88±0.21& 99.09±0.08 &  93.24±0.27 & 92.75±0.27 & 92.98±0.22 & 99.61±0.02 \\
\bottomrule
\end{tabular} }

\vspace{0.3em}

  \scalebox{0.9}{\begin{tabular}{l
  *{2}{>{\centering\arraybackslash}p{1.3cm}
        >{\centering\arraybackslash}p{1.3cm}
        >{\centering\arraybackslash}p{1.3cm}
        >{\centering\arraybackslash}p{1.3cm}}}
\toprule
& \multicolumn{4}{c}{\bfseries OrganSMNIST}
& \multicolumn{4}{c}{\bfseries TissueMNIST} \\
\cmidrule(lr){2-5}\cmidrule(lr){6-9}
\bfseries Method
  & ACC & F1   & Sensitivity  & ROC-AUC
  & ACC & F1   & Sensitivity  & ROC-AUC \\
\midrule
GCN & 78.62±0.82 & 73.74±0.99 & 73.85±0.08 & 97.80±0.11
    & 50.90±0.32 & 32.61±0.79 & 32.51±0.07 & 81.98±0.31\\
\rowcolor{green!40}
GCN + Reasoner & 79.34±1.36 & 74.81±1.51 & 75.23±0.14 & 97.94±0.17 & 51.51±0.36 & 34.30±0.30 & 34.21±0.15 & 82.66±0.11 \\
\arrayrulecolor{gray}\hline
GAT & 81.80±0.68 & 77.22±0.73 & 77.16±0.07 & 98.39±0.09 & \cellcolor{green!40}51.53±0.35 & 33.10±0.56 & 33.06±0.07 & \cellcolor{green!40}83.11±0.12  \\
\rowcolor{green!40}
GAT + Reasoner & 82.08±0.59& 77.70±0.69 & 77.99±0.72 & 98.36±0.09 & \cellcolor{white}43.98±0.41 & 37.59±0.40& 41.02±0.42 & \cellcolor{white}82.51±0.29 \\
\arrayrulecolor{gray}\hline
GIN &  77.23±0.62 & 71.65±0.65 & 71.77±0.76 & 97.36±0.10 & \cellcolor{green!40}50.51±1.09 & 30.31±3.70 & 32.50±2.05 & 81.72±0.85\\
\rowcolor{green!40}
GIN + Reasoner & 80.29±0.60 & 75.70±0.74 & 75.99±0.63 & 98.00±0.10 & \cellcolor{white}43.26±0.54 & 36.91±0.46 & 40.53±0.41& 82.29±0.19 \\

\bottomrule
\end{tabular}}

\vspace{0.3em}

  \scalebox{0.9}{\begin{tabular}{l
  *{2}{>{\centering\arraybackslash}p{1.3cm}
        >{\centering\arraybackslash}p{1.3cm}
        >{\centering\arraybackslash}p{1.3cm}
        >{\centering\arraybackslash}p{1.3cm}}}
\toprule
& \multicolumn{4}{c}{\bfseries BloodMNIST}
& \multicolumn{4}{c}{\bfseries MorphoMNIST} \\
\cmidrule(lr){2-5}\cmidrule(lr){6-9}
\bfseries Method
  & ACC & F1   & Sensitivity  & ROC-AUC
  & ACC & F1   & Sensitivity  & ROC-AUC \\
\midrule
GCN & 80.49±0.66 & 77.46±0.96 & 77.15±0.09 & 96.56±0.18
    & \cellcolor{green!40}90.89±0.05 & 90.82±0.02 & 90.73±0.02 & 98.88±0.03 \\
\rowcolor{green!40}
GCN + Reasoner & 80.32±0.55 & 78.01±0.62 & 78.18±0.07 & 96.77±0.10 & \cellcolor{white}90.78±0.48 & 90.97±0.35 & 90.78±0.53 & 98.90±0.20 \\
\arrayrulecolor{gray}\hline
GAT & 82.02±0.31 & 79.38±0.40 & 79.18±0.04 & 97.45±0.08 & 91.50±0.15 & 91.48±0.16 & 91.33±0.05 & 98.73±0.04 \\
\rowcolor{green!40}
GAT+ Reasoner& 80.59±0.62 & 78.29±0.70 & 79.31±0.56 & 97.18±0.11 & 91.69±0.32 & 91.65±0.20 & 91.66±0.18 & 98.79±0.02 \\
\arrayrulecolor{gray}\hline
GIN &  80.30±0.60 & 77.21±0.91 & 76.44±0.95 & 96.58±0.21 & \cellcolor{green!40}91.60±0.19 & 91.59±0.18 & 91.60±0.19 & \cellcolor{green!40}98.75±0.05\\
\rowcolor{green!40}
GIN + Reasoner & 80.56±0.37 & 78.19±0.35 & 79.18±0.32 & 96.89±0.07 & \cellcolor{white}91.55±0.85 & 91.64±0.55 & 91.63±0.26 & \cellcolor{white}98.74±0.06 \\
\bottomrule
\end{tabular}}
\vspace{-6mm}
\end{table} 

\noindent \textbf{Classification performance}
X-Node consistently improves over baseline GNNs across all datasets. For example, on OrganAMNIST, it raises F1 from 91.19\% to 93.16\% and sensitivity from 91.18\% to 94.07\% as in Table \ref{gnn_varaints}. Gains are especially notable in sensitivity—crucial in medical diagnosis showing up to 3–5\% improvement. On large-scale graphs like TissueMNIST and MorphoMNIST, although overall accuracy slightly drops, X-Node improves F1 and sensitivity metrics which is critical in medical domain.

\noindent \textbf{Interpretability (Per-Node Explanations)} Beyond classification accuracy, X-Node generates faithful, introspective justifications for each node’s prediction. For example, Node 3 from Fig.~\ref{fig:niyan}, which produces the following self-explanation based on its context vector $\mathbf{c}_i$ and prediction $\hat{y}_i$ supporting \textbf{H1}. This reasoning aligns closely with graph-theoretic descriptors like neighborhood sparsity, edge strength, and label agreement:
\vspace{-1mm}
\definecolor{lightpastelmintgreen}{RGB}{230, 255, 240} 


\begin{tcolorbox}[
  colback=lightpastelmintgreen,
  colframe=teal!60!black,
  title={Node 3 (True = \texttt{Kidney-Right}, Pred = \texttt{Femur-Left}) Explanation},
  boxrule=0.6pt,
  arc=4pt
]
\vspace{-1mm}
\footnotesize
\textit{“I predicted that I was representing the 'femur-left' class, but my true label is 'kidney-right'. Let's reflect on this. I have a degree of 4, meaning I'm moderately connected. My clustering coefficient is 0.00, which shows my neighbors are not well connected to each other. One potential cause for the error could be my high average edge weight (0.929), indicating strong connections that might have biased my prediction. While my top feature (F[117] = 10.00) may be distinctive for kidney-right, it was likely overridden by misleading neighborhood influence. My community label and low centrality further suggest I’m not centrally embedded in the kidney-right cluster. This misclassification reveals how structural signals can dominate node identity when feature signals are ambiguous.”}
\end{tcolorbox}

\noindent The explanation is both structured and contextualized, allowing users to \textbf{trace misclassification} as in this example back to interpretable topological features. This per-node transparency highlights the value of explanation-aware training.

\noindent \textbf{Conclusion} This work introduces \textbf{X-Node}, a self-explainable GNN architecture that integrates per-node explanation into training. Beyond accuracy, it enables interpretable decision-making through learned context vectors and LLM-based textual rationales. \textit{Grok 3} outperforms \textit{Gemini 2.5 Pro} in clarity of explanation, and future work can explore the effect of prompt variations. X-Node offers a transferable framework for faithful, explanation-aware learning marking a step toward trustworthy graph intelligence.

%
%
%
%

\bibliographystyle{splncs04}
\bibliography{GRAIL2025-paper11/bibliography}

\begin{thebibliography}{10}
\providecommand{\url}[1]{\texttt{#1}}
\providecommand{\urlprefix}{URL }
\providecommand{\doi}[1]{https://doi.org/#1}

\bibitem{graph_med_diagnosis}
Ahmedt-Aristizabal, D., Armin, M.A., Denman, S., Fookes, C., Petersson, L.: Graph-based deep learning for medical diagnosis and analysis: Past, present and future. Sensors  \textbf{21}(14), ~4758 (Jul 2021). \doi{10.3390/s21144758}

\bibitem{self_explaining_nn}
Alvarez~Melis, D., Jaakkola, T.: Towards robust interpretability with self-explaining neural networks. In: Bengio, S., Wallach, H., Larochelle, H., Grauman, K., Cesa-Bianchi, N., Garnett, R. (eds.) Advances in Neural Information Processing Systems. vol.~31. Curran Associates, Inc. (2018), \url{https://proceedings.neurips.cc/paper_files/paper/2018/file/3e9f0fc9b2f89e043bc6233994dfcf76-Paper.pdf}

\bibitem{post_hoc_fail}
Bordt, S., Finck, M., Raidl, E., von Luxburg, U.: Post-hoc explanations fail to achieve their purpose in adversarial contexts. In: Proceedings of the 2022 ACM Conference on Fairness, Accountability, and Transparency. p. 891–905. FAccT '22, Association for Computing Machinery, New York, NY, USA (2022). \doi{10.1145/3531146.3533153}, \url{https://doi.org/10.1145/3531146.3533153}

\bibitem{going_beyond_euclidian}
Bronstein, M.M., Bruna, J., LeCun, Y., Szlam, A., Vandergheynst, P.: Geometric deep learning: Going beyond euclidean data. IEEE Signal Processing Magazine  \textbf{34}(4),  18--42 (2017). \doi{10.1109/MSP.2017.2693418}

\bibitem{moprhomnist_data}
Coelho~de Castro, D., Tan, J., Kainz, B., Glocker, B.: Morpho-mnist: Quantitative assessment and diagnostics for representation learning. Journal of Machine Learning Research  \textbf{20} (10 2019)

\bibitem{Gxain}
Cedro, M., Martens, D.: Graphxain: Narratives to explain graph neural networks (11 2024). \doi{10.48550/arXiv.2411.02540}

\bibitem{GNN_Neural}
Gao, Y., Yang, H., Chen, Y., Wu, J., Zhang, P., Wang, H.: Llm4gnas: A large language model based toolkit for graph neural architecture search (02 2025). \doi{10.48550/arXiv.2502.10459}

\bibitem{whyXAI_medical_domain}
Holzinger, A., Biemann, C., Pattichis, C., Kell, D.: What do we need to build explainable ai systems for the medical domain?  (12 2017). \doi{10.48550/arXiv.1712.09923}

\bibitem{faithful_interpretable}
Jacovi, A., Goldberg, Y.: Towards faithfully interpretable {NLP} systems: How should we define and evaluate faithfulness? In: Jurafsky, D., Chai, J., Schluter, N., Tetreault, J. (eds.) Proceedings of the 58th Annual Meeting of the Association for Computational Linguistics. pp. 4198--4205. Association for Computational Linguistics, Online (Jul 2020). \doi{10.18653/v1/2020.acl-main.386}, \url{https://aclanthology.org/2020.acl-main.386/}

\bibitem{kipf2017semisupervised}
Kipf, T.N., Welling, M.: Semi-supervised classification with graph convolutional networks. In: International Conference on Learning Representations (2017), \url{https://openreview.net/forum?id=SJU4ayYgl}

\bibitem{gated_Fully_fusion}
Li, X., Zhao, H., Han, L., Tong, Y., Tan, S., Yang, K.: Gated fully fusion for semantic segmentation. Proceedings of the AAAI Conference on Artificial Intelligence  \textbf{34},  11418--11425 (04 2020). \doi{10.1609/aaai.v34i07.6805}

\bibitem{egnn_fire}
Li, Z., Geisler, S., Wang, Y., Günnemann, S., Leeuwen, M.: Explainable graph neural networks under fire (06 2024). \doi{10.48550/arXiv.2406.06417}

\bibitem{parameterized_explainer}
Luo, D., Cheng, W., Xu, D., Yu, W., Zong, B., Chen, H., Zhang, X.: Parameterized explainer for graph neural network. In: Larochelle, H., Ranzato, M., Hadsell, R., Balcan, M., Lin, H. (eds.) Advances in Neural Information Processing Systems. vol.~33, pp. 19620--19631. Curran Associates, Inc. (2020), \url{https://proceedings.neurips.cc/paper_files/paper/2020/file/e37b08dd3015330dcbb5d6663667b8b8-Paper.pdf}

\bibitem{PGExplainar}
Luo, D., Cheng, W., Xu, D., Yu, W., Zong, B., Chen, H., Zhang, X.: Parameterized explainer for graph neural network. In: Proceedings of the 34th International Conference on Neural Information Processing Systems. NIPS '20, Curran Associates Inc., Red Hook, NY, USA (2020)

\bibitem{disease_pred_gnn}
Parisot, S., Ktena, S.I., Lee, M., Guerrero, R., Glocker, B., Rueckert, D.: Disease prediction using graph convolutional networks: Application to autism spectrum disorder and alzheimer’s disease. Medical Image Analysis  \textbf{48} (06 2018). \doi{10.1016/j.media.2018.06.001}

\bibitem{xai_deeplearning}
Samek, W., Wiegand, T., Müller, K.R.: Explainable artificial intelligence: Understanding, visualizing and interpreting deep learning models. ITU Journal: ICT Discoveries - Special Issue 1 - The Impact of Artificial Intelligence (AI) on Communication Networks and Services  \textbf{1},  1--10 (10 2017). \doi{10.48550/arXiv.1708.08296}

\bibitem{LIMESHAP}
Slack, D., Hilgard, S., Jia, E., Singh, S., Lakkaraju, H.: Fooling lime and shap: Adversarial attacks on post hoc explanation methods. In: Proceedings of the AAAI/ACM Conference on AI, Ethics, and Society. p. 180–186. AIES '20, Association for Computing Machinery, New York, NY, USA (2020). \doi{10.1145/3375627.3375830}, \url{https://doi.org/10.1145/3375627.3375830}

\bibitem{gat}
Veličković, P., Cucurull, G., Casanova, A., Romero, A., Liò, P., Bengio, Y.: Graph attention networks. In: International Conference on Learning Representations (2018), \url{https://openreview.net/forum?id=rJXMpikCZ}

\bibitem{chainthought_llm}
Wei, J., Wang, X., Schuurmans, D., Bosma, M., Ichter, B., Xia, F., Chi, E.H., Le, Q.V., Zhou, D.: Chain-of-thought prompting elicits reasoning in large language models. In: Proceedings of the 36th International Conference on Neural Information Processing Systems. NIPS '22, Curran Associates Inc., Red Hook, NY, USA (2022)

\bibitem{gin_powerfulGNN}
Xu, K., Hu, W., Leskovec, J., Jegelka, S.: How powerful are graph neural networks? In: International Conference on Learning Representations (2019), \url{https://openreview.net/forum?id=ryGs6iA5Km}

\bibitem{yang2023medmnist}
Yang, J., Shi, R., Wei, D., et~al.: Medmnist v2 - a large-scale lightweight benchmark for 2d and 3d biomedical image classification. Scientific Data  \textbf{10}, ~41 (2023). \doi{10.1038/s41597-022-01721-8}, \url{https://doi.org/10.1038/s41597-022-01721-8}

\bibitem{gnn_explainer}
Ying, Z., Bourgeois, D., You, J., Zitnik, M., Leskovec, J.: Gnnexplainer: Generating explanations for graph neural networks. In: Wallach, H., Larochelle, H., Beygelzimer, A., d\textquotesingle Alch\'{e}-Buc, F., Fox, E., Garnett, R. (eds.) Advances in Neural Information Processing Systems. vol.~32. Curran Associates, Inc. (2019), \url{https://proceedings.neurips.cc/paper_files/paper/2019/file/d80b7040b773199015de6d3b4293c8ff-Paper.pdf}

\bibitem{XGNN}
Yuan, H., Tang, J., Hu, X., Ji, S.: Xgnn: Towards model-level explanations of graph neural networks. pp. 430--438 (08 2020). \doi{10.1145/3394486.3403085}

\bibitem{LLMExplainer}
Zhang, J., Liu, J., Luo, D., Neville, J., Wei, H.: Llmexplainer: Large language model based bayesian inference for graph explanation generation (07 2024). \doi{10.48550/arXiv.2407.15351}

\end{thebibliography}

\end{document}